# Point2Building: Reconstructing buildings from airborne LiDAR point clouds

Yujia Liu [a], Anton Obukhov [a], Jan Dirk Wegner [b], Konrad Schindler [a,*]

[a] *Photogrammetry and Remote Sensing, ETH Zürich, 8093 Zürich, Switzerland*
[b] *Department of Mathematical Modeling and Machine Learning, University of Zürich, 8057 Zürich, Switzerland*

A B S T R A C T

We present a learning-based approach to reconstructing buildings as 3D polygonal meshes from airborne LiDAR point clouds. What makes 3D building reconstruction from airborne LiDAR difficult is the large diversity of building designs, especially roof shapes, the low and varying point density across the scene, and the often incomplete coverage of building facades due to occlusions by vegetation or the sensor's viewing angle. To cope with the diversity of shapes and inhomogeneous and incomplete object coverage, we introduce a generative model that directly predicts 3D polygonal meshes from input point clouds. Our autoregressive model, called Point2Building, iteratively builds up the mesh by generating sequences of vertices and faces. This approach enables our model to adapt flexibly to diverse geometries and building structures. Unlike many existing methods that rely heavily on pre-processing steps like exhaustive plane detection, our model learns directly from the point cloud data, thereby reducing error propagation and increasing the fidelity of the reconstruction. We experimentally validate our method on a collection of airborne LiDAR data from Zurich, Berlin, and Tallinn. Our method shows good generalization to diverse urban styles.

## 1. Introduction

The conversion of urban environments into digital representations is important across remote sensing, computer vision, computer graphics, and photogrammetry. Three-dimensional models of urban structures offer broad utility in applications such as visualization, navigation, and urban planning (Poullis and You, 2009; Lafarge and Mallet, 2012; Cheng et al., 2011). Among sensing technologies for urban modeling, airborne Light Detection and Ranging (LiDAR) (Liu, 2008; Chen, 2007; Li et al., 2020a; Sampath and Shan, 2007) has become particularly important. Although point clouds from LiDAR are just one of many data sources for 3D modeling, their ability to accurately capture large-scale geometry is particularly promising for city-scale perception of urban environments. Various formats were developed to represent 3D surfaces. Examples include NURBS for smooth, accurate surfaces, voxel grids for volumetric analysis, subdivision surfaces for computer graphics and animation, and implicit surfaces for mathematical descriptions or scientific visualizations. Converting point clouds to such explicit or implicit representations involves diverse methods, from traditional geometry-based techniques (Hoppe et al., 1992; Kazhdan et al., 2006; Boissonnat and Oudot, 2005) to learning-based approaches (Groueix et al., 2018; Guerrero et al., 2018; Park et al., 2019). Methods for explicit surfaces utilize meshes (Gupta, 2020; Jiang et al., 2020, 2019) or patches (Badki et al., 2020; Groueix et al., 2018; Ma et al., 2021) to directly define geometry, known for their efficiency. Implicit methods represent shapes with functions over voxel grids (Dai et al., 2018; Jiang et al., 2017; Liao et al., 2018) or neural networks (Chen and Zhang, 2019; Mescheder et al., 2019; Park et al., 2019; Liu et al., 2023b), offering a high level of visual detail and compelling view synthesis (Martin-Brualla et al., 2021; Mildenhall et al., 2021), but at a higher computational cost.

Among these surface representations, we focus on 3D polygonal meshes representing buildings. Polygonal meshes (Botsch et al., 2010; Chen et al., 2020; Nan and Wonka, 2017) are an effective representation format comprising interconnected vertices and faces. It provides compact structural information beyond dense triangulations (Fabio et al., 2003; Salman et al., 2010) and thus appears an apt choice for the 3D representation of urban areas.

Traditional approaches to 3D polygon reconstruction (Halla, 1999; Verma et al., 2006) have relied on learning-free algorithms that are often rigid and have limited adaptability. Furthermore, a few existing learning-based methods (Chen et al., 2022, 2023a) often suffer from a lack of flexibility, being tied down by the constraints of their initial processing stages, like primitive segmentation (Liu et al., 2023b) or planar segment detection (Yang et al., 2010) with methods like RANSAC (Schnabel et al., 2007; Li and Shan, 2022). Roof reconstruction, in particular, has received significant attention due to its important role in urban mapping. Roof plane segmentation is often based on variants of region growing that, starting from seed points, attempt to flood-fill roof segments based on geometric criteria (e.g., Gorte (2002), Vo et al. (2015), Cao et al. (2017), Xu et al. (2017), Gilani et al. (2018)). The approach is straightforward but sensitive to how seed points are selected; it also struggles with accurately determining segment boundaries. To overcome those limitations, energy minimization methods have been introduced. Yan et al. (2014), Dong et al. (2018), Wang and Ji (2021) treat roof plane refinement as a multi-label optimization problem and solve it with the graph cut algorithm. Since this is comparatively expensive in terms of computation, Li et al. (2020b), Liu et al. (2023a) improve efficiency with boundary relabeling techniques that refine an initial set of planar patches. Feature clustering methods describe points in a geometrically meaningful feature space and group them into planar patches by clustering in that space. Filin and Pfeifer (2006) used a slope-adaptive neighborhood system to calculate the point features precisely. Sampath and Shan (2009) employ fuzzy $k$-means based on point normals. Zhang et al. (2018) use spectral clustering of straight line segments to detect roof planes. Chen et al. (2023b) introduced a tangent plane distance metric and used adaptive DBSCAN to cluster coplanar points. The main challenge with feature clustering is the design of features capable of reliably distinguishing plane instances. Handcrafted features often result in inaccurate boundaries and over-segmentation. To mitigate this, Zhang and Fan (2022), Li et al. (2023) used deep features and designed a multi-task network that embeds point clouds into a high-dimensional space so that points on the same plane are assigned similar embeddings. Still, that method also struggles to delineate boundaries accurately. An issue shared by methods that first segment the point cloud, then reconstruct the faces is that mistakes in the first step propagate and cause errors in the final model. Building upon these insights, we aim to employ a more integrated learning-based approach, one that avoids the dependency on initial segmentation or plane detection and reduces the risk of error accumulation.

To this end, we adopt an autoregressive generative model, PolyGen (Nash et al., 2020), to reconstruct polygonal buildings from 3D LiDAR point clouds. PolyGen is especially well-suited for our needs because it naturally handles the diversity in building topology without relying on pre-segmentation or prior primitive fitting. It brings the advantage of learning directly from data, allowing for a more effective and adaptive representation of architectural features than existing methods reliant on hand-designed geometric descriptors. PolyGen operates in two stages. First, it produces the positions of vertices based on the input 3D point cloud data, effectively capturing the spatial configuration of the target structures. Second, it constructs the mesh faces that connect the created vertices, forming the object surface.

The autoregressive nature of PolyGen's generative process is affected by common issues shared by most modern generative models (e.g., hallucinations). We turn the built-in diversity of the generated output into an advantage by sampling multiple reconstructions conditioned on the same input point cloud and keeping only the one that best satisfies a set of plausibility checks.

PolyGen is primarily designed for unconditioned mesh generation. A proof of concept for conditional generation is limited to gridded inputs such as images and voxels. Here, we design an encoder that operates on arbitrary 3D points to enable the processing of geospatial point clouds. Moreover, PolyGen (and other learned, generative models) sometimes produce geometrically implausible results. Hence, we interleave the generation process with validity checks that reject implausible hypotheses early and rerun the appropriate generation steps. These extensions make it possible to generate meaningful polygon meshes in the large majority of cases.

We have conducted experiments on a combined dataset encompassing aerial LiDAR scans sourced from the Federal Office of Topography (swisstopo, 2023) and corresponding 3D polygonal mesh models provided by the City of Zurich (Stadt Zurich, 2023). Our method trained on this dataset shows improvements compared to existing approaches across various evaluation metrics. Additionally, to demonstrate the method's adaptability and effectiveness, we include experiments on further datasets from Berlin and Tallinn in our evaluation.

## 2. Related work

Reconstructing buildings from point cloud data (Brenner et al., 2001) involves specialized techniques that address the regular and deliberate design typical of architectural structures. These methods are developed to accurately represent essential architectural features such as flat surfaces, sharp edges, and key elements like floorplans, facades, and outer walls. They must also be flexible enough to handle buildings of various sizes and complexities, from small single-family homes to large commercial complexes. In reviewing the field of building geometry reconstruction, our focus is primarily on the generation of building models in the polygonal representation. The methods in this area can be broadly categorized into model-based, graph-based, and space-partitioning-based approaches.

### 2.1. Model-based approaches

Model-based (a.k.a. template-based) methods are common in 3D building reconstruction from point clouds. These methods operate under the assumption that the complex architecture of buildings can be approximated using a library of geometric standard shapes. For example, the Manhattan-world assumption limits building surface orientations to three primary directions, representing buildings with axis-aligned polycubes (Li et al., 2016a,b). Generally speaking, building roofs have received the most attention. Roofs are modeled as combinations of predefined shapes (Kada and McKinley, 2009; Huang et al., 2013) like a saddleback, pent, flat, tent, mansard, etc. These methods typically start with extracting and refining the outlines of building footprints. The refined footprints are segmented into distinct 2D sections. The segmentation subdivides the reconstruction task into smaller subtasks, each identifying a roof style and fitting the corresponding model parameters, steps for which established methods are available. For example, Poullis et al. (2008) uses Gaussian mixture models to determine the parameters of basic shapes. Kada and McKinley (2009) rely on the orientation of roof surfaces to classify roof types and employ specific height measurements for parameter estimation. Xu et al. (2015) propose an improved, weighted RANSAC algorithm for roof plane segmentation, factoring in point-to-plane distance and normal vector differences. Li et al. (2017) extend RANSAC with normal distribution transformation cells as sampling units to obtain better fits. Dal Poz and Yano Ywata (2020) develop an adaptive RANSAC algorithm that adjusts distance and angle thresholds for consistency with the hypothesized plane model. Techniques like the reversible jump Markov Chain Monte Carlo model (Huang et al., 2013) have also been utilized to fit roof parameters to the input data. Xiong et al. (2014) design a graph edit dictionary to correct errors in roof topology graphs reconstructed from point clouds. Xiong et al. (2015) proposed flexible building primitives for 3D building modeling, enhancing the adaptability of the models. While model-based methods provide a natural framework for reconstructing 3D buildings, they are bounded by their inherent assumptions, especially the selected collection of templates, which may not be able to cover the full architectural diversity. Recent advances in urban modeling have increasingly been driven by techniques that offer greater flexibility and can move beyond the constraints of predefined template shapes.

### 2.2. Graph-based approaches

Various graph-based methods have also been investigated (Chen and Chen, 2008; Schindler et al., 2011; Van Kreveld et al., 2011; Sampath and Shan, 2009). They generally focus on extracting geometric primitives, which are then linked into a graph that captures the adjacency relations. By intersecting these planar primitives, edges are reconstructed, forming a building model (Dorninger and Pfeifer, 2008; Maas and Vosselman, 1999; Vosselman et al., 2001). Challenges emerge when missing or spurious connections between primitives disrupt the graph's connectivity, degrading the reconstruction quality. To address these difficulties, a hybrid approach (Lafarge and Alliez, 2013; Labatut et al., 2009) was introduced that combines two distinct elements: it uses polyhedral shapes to mark regions with high confidence levels and dense triangle meshes to represent areas of greater complexity accurately. Thus, it balances precision and the ability to handle complex structures. Another method (Zhou and Neumann, 2010, 2011), 2.5D dual contouring. It starts by projecting the point cloud data into a two-dimensional grid. Next, it finds so-called hyper-points within this grid, i.e., sets of 3D points having the same $x$-$y$ coordinates yet different $z$ values. These are then connected to form a graph optimized by collapsing subtrees and merging similar branches to reduce complexity. Finally, a seamless mesh is generated by connecting the hyper-points into a watertight model. Wu et al. (2017) have developed a graph-based method for reconstructing 3D building models from airborne LiDAR point clouds, focusing on efficient and accurate model generation. Nys et al. (2020) present an automatic method for compact 3D building reconstruction from LiDAR point clouds, emphasizing data reduction and model precision. To better handle complex shapes, Song et al. (2020) propose a technique for reconstructing curved buildings from airborne LiDAR data by matching and deforming geometric primitives. Yang et al. (2022) introduce the connectivity-aware graph, a planar topology representation for 3D building surfaces that enhances connectivity and structural accuracy.

Graph-based methods share the advantage that they are not confined to a predetermined set of geometric primitives, allowing for the reconstruction of various building structures. Nonetheless, these techniques may encounter issues related to topological correctness and can be computationally demanding.

*2.3. Space-partition-based approaches*

Another class of building reconstruction methods involves dividing the volume containing the input point cloud into a redundant set of elementary shapes, which are then pruned to form an approximate building model. For example, Verdie et al. (2015) breaks down the space around the building into polyhedral cells, which are then evaluated by a Min-Cut formulation to determine whether they fall within the building polyhedra. Chauve et al. (2010) propose a robust, piecewise planar 3D reconstruction method that effectively handles large-scale unstructured point data. PolyFit (Nan and Wonka, 2017) generates a collection of potential faces by the intersecting planes identified with RANSAC. Selecting the faces that most accurately represent the building's surface is approached as a binary integer linear program with a strict constraint of edge-face connections. PolyFit is computationally intensive, particularly when handling complex structures that yield many potential faces to be considered. Bauchet and Lafarge (2020) introduce a kinetic shape reconstruction framework that generates lighter decompositions of space into primitives based on a kinetic data structure. Xie et al. (2021) propose to combine rule-based and hypothesis-based strategies for efficient building reconstruction. Li and Wu (2021) have extended PolyFit for building reconstruction from photogrammetric point clouds to better leverage the relations between primitives for procedural modeling. Huang et al. (2022) also extend PolyFit for airborne LiDAR point clouds. They introduce a new energy term that captures shape preferences for roofs and add two hard constraints that ensure correct topology and enhance the recovery of small details. Several works are based on adaptive space partitioning in combination with deep learning. Chen et al. (2022) propose a method for reconstructing compact building models from point clouds with deep implicit fields to achieve high-fidelity models. Their approach leverages deep learning to implicitly represent building surfaces, resulting in more accurate and compact models. Chen et al. (2023a) introduce PolyGNN, a polyhedron-based graph neural network for 3D building reconstruction from point clouds. That method uses graph neural networks to model the complex relationships between building components with enhanced robustness.

Another line of works focuses on geometric simplification to obtain compact surfaces by simplifying dense triangle meshes. Salinas et al. (2015) introduce a structure-aware mesh decimation method that maintains geometric fidelity while reducing mesh complexity. Bouzas et al. (2020) presents a structure-aware approach for mesh simplification that better preserves structural details. Li and Nan (2021) have developed a feature-preserving 3D mesh simplification technique for urban buildings that retains important architectural features while reducing the overall mesh size. All these methods start from complete but redundant meshes and aim to balance preserving detail against compactness and efficiency. In addition to mesh simplification, several recent works recover building shape through wireframe reconstruction. Huang et al. (2023) propose a method to reconstruct detailed wireframes from LiDAR data. Luo et al. (2022) construct 3D building wireframes from 3D line clouds with a deep learning approach. Jiang et al. (2023) presents a technique to extract 3D lines of buildings from ALS point clouds with a graph neural network that embeds information about building corners to recover structurally important edges. Li et al. (2022) creatively divide roof wireframe extraction task into vertex and edge prediction problems, and propose an end-to-end neural network to directly predict vertices and edges of roof models from point cloud data. Their experimental results demonstrate that deep networks can be successfully applied for roof model reconstruction.

Upon review, it becomes evident that each 3D reconstruction approach has different advantages and limitations. Model-based methods are structured and efficient, suitable for quick reconstruction tasks. However, they often depend on a finite set of predefined shapes, which may limit their applicability to diverse architectural styles. Graph-based strategies offer greater flexibility and adaptability to handle complex structures yet tend to be prone to connectivity errors affecting the reconstructed models' integrity. Conversely, space-partition-based techniques provide a detailed decomposition of 3D space but face challenges in computational efficiency, particularly when dealing with large-scale structures.

**3. Method**

We use an autoregressive model to predict mesh elements conditioned on the input point cloud sequentially. This method is distinct from conventional 3D reconstruction techniques, relying on a sequential, generative process rather than discriminative classification or regression of shape primitives (respectively, graphs). It is inspired by and adapted from PolyGen (Nash et al., 2020), which pioneered autoregressive transformer models (Vaswani et al., 2017) for 3D mesh generation.

The method consists of two main modules, as shown in Fig. 1: the vertex generation module comprising a point cloud feature extractor and the face module. The vertex module operates by predicting the positions of vertices conditioned on point cloud features, effectively learning the underlying geometric structure of the shapes it is trained on. After vertex prediction, the face module sequentially generates mesh faces that connect the vertices to form the object surface. Each generated mesh is evaluated against predefined plausibility rules, ensuring the final model meets minimal criteria for a valid building model. The process is repeated until the mesh meets all required criteria.

*3.1. Vertex module*

The vertex module integrates a point cloud feature extractor with a transformer-based decoder and generates a joint probability distribution for a sequence of vertices, as depicted in Fig. 2.

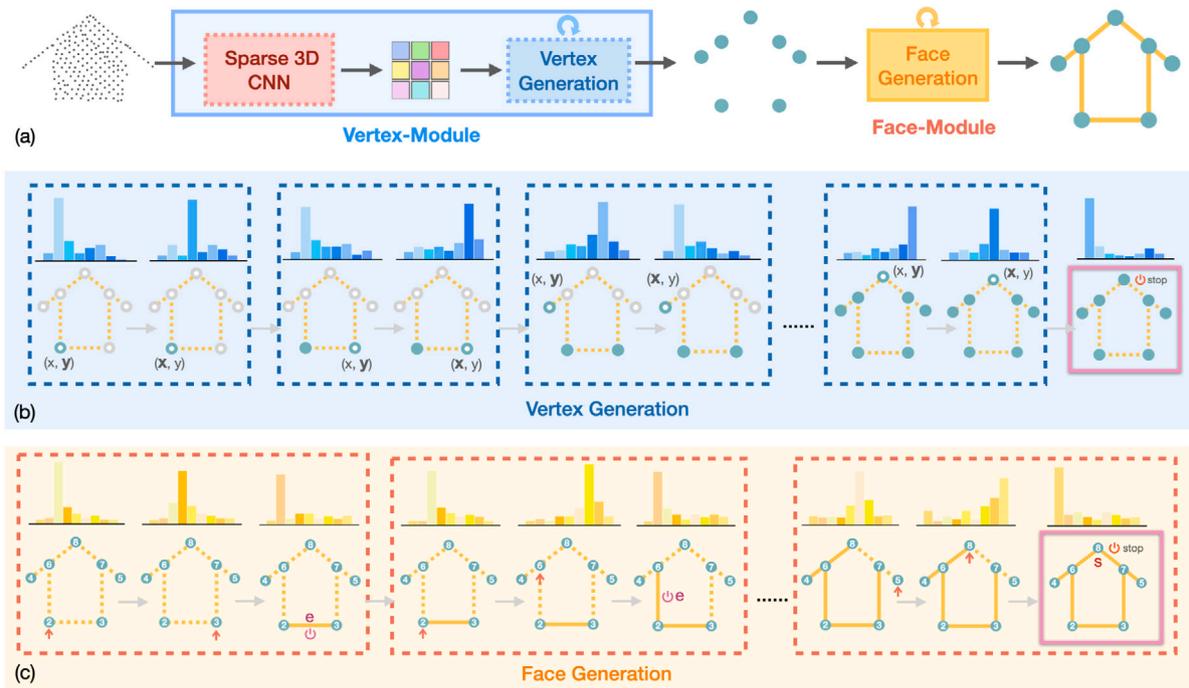

**Fig. 1.** (a) Overview of the proposed pipeline visualized with cartoon 2D data. The actual implementation operates in 3D. (b) Stages of data processing. Each stage generates tuples of coordinates of one vertex. In the given 2D case, $(y, x)$ tuples are generated. For real data, we generate $(z, y, x)$ tuples. (c) Stages of assembling vertices into surfaces. At each stage, only one surface primitive corresponding to one line segment in the visualized cartoon 2D case) or a polyline capturing a polygon on a single plane (in the real 3D case) is created.

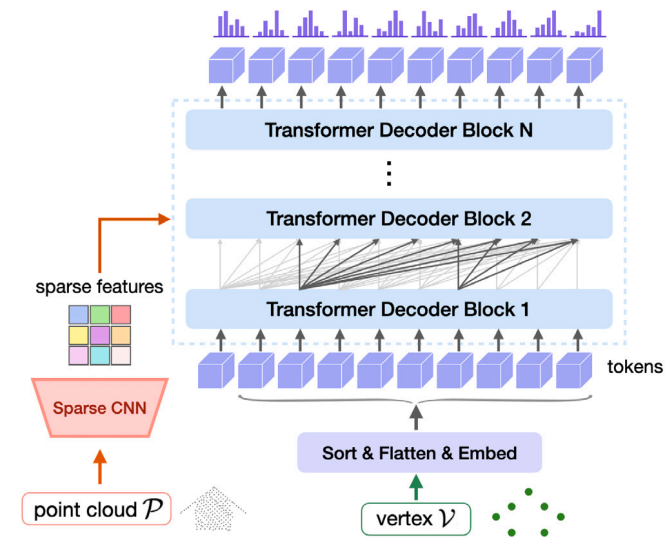

**Fig. 2.** Illustration of vertex module architecture, comprising a point cloud encoder and a transformer-based decoder. Initially, the input point cloud is processed through SparseCNN, resulting in a sequence of context embeddings. The transformer decoder then applies cross-attention to these embeddings following a self-attention layer, analogous to the original machine translation Transformer model. Vertices are ordered according to their $z$, $y$, and $x$ coordinates, converted into a flat sequence, and subsequently transformed into tokens as detailed in Section 3.1.2. For each token, the module predicts a distribution for the subsequent vertex coordinate value.

### 3.1.1. Sparse 3D CNN for feature extraction

The input point cloud (normalized to fit within a unit sphere) is encoded using a sparse convolutional neural network (SparseCNN), as implemented in the Minkowski Engine (Choy et al., 2019). This network takes a set of 3D points, denoted as $\mathcal{P}$, and outputs a set of context embeddings (Guo et al., 2022).

Initially, these points are discretized using a grid with a resolution of $128 \times 128 \times 128$. Subsequent stages involve convolutional operations paired with pooling to gradually reduce the spatial resolution of the data to a $16 \times 16 \times 16$ grid, stored in the form of grid indices and feature vectors.

### 3.1.2. Block for vertex generation

The second block learns to produce vertex sequences conditioned on voxelized features extracted in the previous step.

**Quantization-Driven Vertex Modeling.** The vertices of a mesh are treated as discrete units through quantization. By implementing an 8-bit uniform quantization within the bounding box of the point cloud, we compress the point cloud information, merging vertices that are close to each other, akin to how SparseCNN discretizes space into a grid. We model the vertices using a categorical distribution along each axis. The generative phase of our model outputs logits that align with this distribution. The usage of quantization is similar to that applied by van den Oord et al. (2016a,b), leveraging a categorical distribution for its flexibility in modeling various distributions without making assumptions about their shape.

**Vertex Sequence Modeling.** The vertex module outputs a probability distribution over vertex coordinates conditioned on previously predicted vertex coordinates and input features at each step. Sequence element order and other sequence formation conventions are important to any autoregressive framework. Therefore, ground truth vertex coordinate tuples $(x, y, z)$ are reversed and sorted lexicographically to establish a reliable vertex order from bottom to top, left to right, and far to near. Subsequently, the list of sorted tuples is flattened into a raw sequence of tokens $(z_0, y_0, x_0, z_1, y_1, x_1, z_2, y_2, x_2, \ldots)$, terminated by a stopping token $s$ as shown in Fig. 3.

**Vertex Embedding.** Each input token (quantized vertex coordinate) is encoded with three distinct embeddings: the coordinate embedding, which classifies the token as a spatial coordinate ($x$, $y$ or $z$); the position embedding, which identifies the token's sequential order within the vertex sequence; and the value embedding, which conveys the quantized

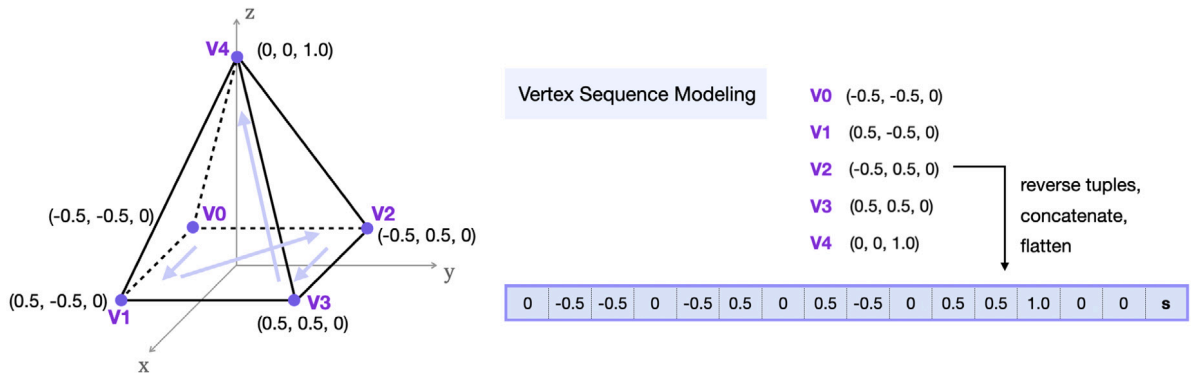

**Fig. 3.** Illustration of creating ordered vertex sequence. The order is established by sorting according to *z*-coordinates, the vertical dimension. If *z* matches, sorting proceeds by *y*, then by *x* to avoid ambiguities. A stopping token *s* is added to signify the conclusion of the flattened vertex sequence.

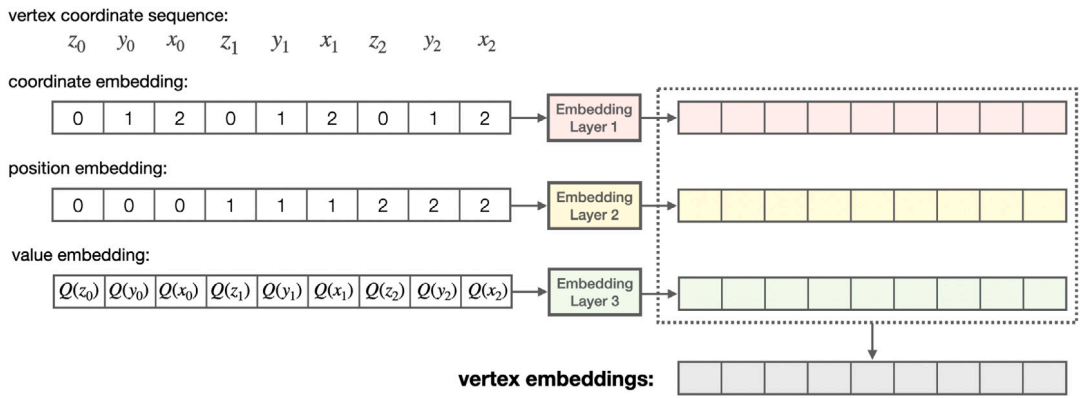

**Fig. 4.** Illustration of vertex embedding process. *x*, *y*, and *z* are coordinates of vertices. $Q(\cdot)$ is the simple quantization function. For each token, the result vertex embeddings are obtained by summing three distinct learned embeddings: the coordinate embedding, the position embedding, and the value embedding. Specifically, Embedding Layer 3 is configured to support a total of 257 unique embeddings, reflecting 256 possible coordinate values plus one additional embedding for the end of a sequence. Each of these embeddings is represented by a vector of 256 dimensions. The coordinate Embedding Layer 1, provides 3 unique embeddings corresponding to the *x*, *y*, and *z* coordinates, while the position embedding layer (Embedding Layer 2) accommodates embeddings for up to a maximum number of vertices +1, each reflecting a distinct position in the sequence.

numerical value of the token's coordinate. The process is demonstrated in Fig. 4.

**Vertex Decoder.** The vertex model employs the Transformer-based decoder introduced by Vaswani et al. (2017). It outputs a joint probability distribution over sequences of vertices. This is achieved by modeling it as a product of conditional distributions, where the likelihood of each vertex coordinate $v_n$ depends on the previous vertex coordinates in the sequence.

$$p(V_{\text{seq}}; \theta) = \prod_{n=1}^{N_v} p(v_n | v_{<n}; \theta)$$

An autoregressive network is used as the backbone for the practical modeling of this distribution in a factorized manner. The model is trained to maximize the likelihood of sampled sequences under the model parameters $\theta$ and input conditions $\mathcal{P}$.

### 3.2. Face module

The Face module architecture employs a transformer-based encoder and decoder, detailed in Fig. 5. It learns to produce a sequence of faces conditioned on the vertices generated by the Vertex Module.

**Face Sequence Modeling.** To structure the mesh faces, we sort them based on vertex indices. This sorting adheres to a simple rule: vertex indices within each face tuple do not decrease. On the sequence level, face tuples are sorted lexicographically. To flatten the list of sorted face tuples, we introduce an 'end-face' token *e* as a delimiter between subsequent tuples. As for the vertex sequences, we terminate the concatenated sequence of tokens with a stopping token *s*, as shown in Fig. 6.

**Face Embeddings.** Similarly to the vertex model, we employ learned embeddings to represent the position and the value associated with each token in the face module. The position of a token is characterized by two aspects: the specific face of the mesh to which it is assigned and its relative position within that face. For the value associated with each token, we assign vertex embeddings generated by a transformer encoder. This embedding process is demonstrated in Fig. 7.

**Face Decoder.** To address the variability in the size of vertex sets across different models, we adopt a strategy akin to the pointer network framework (Vinyals et al., 2015), as employed in PolyGen. This framework starts by encoding the vertex set and then uses an autoregressive network to generate a pointer vector at each step. This pointer vector is compared against the encoded vertex set through a dot-product operation, and the resulting values are then converted into a probability distribution over the vertices by applying the *softmax* function, forming the next face of the mesh.

### 3.3. Mesh generation

We employ multiple strategies for mesh generation during inference. First, we use nucleus sampling to mitigate sample degradation. Second, when generating every single sequence during inference, we define token-level constraints to ensure a valid and meaningful token sequence. Third, once the full sequence has been generated, we perform a semantically informed validation, checking the logical structure and geometric consistency of elements such as floors and walls.

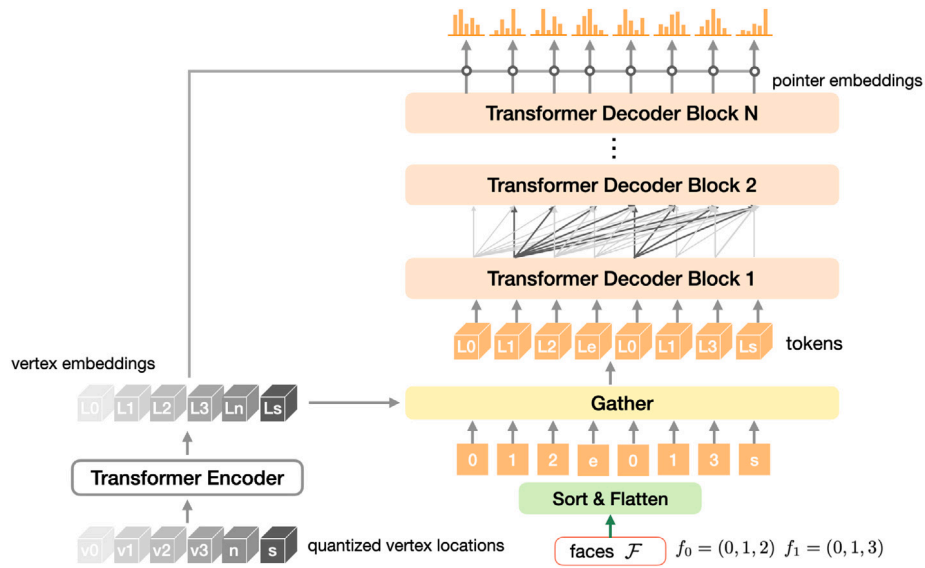

**Fig. 5.** Illustration of the face module architecture, featuring a transformer-based encoder and decoder. The process begins with the transformation of quantized vertex locations by integrating embeddings across the three dimensions (x, y, z), akin to the "value embedding" phase in Fig. 7. These transformed vertex embeddings are further refined by a transformer encoder. For the sorted, flattened faces, a "gather" operation collects the corresponding embeddings for each vertex index within the faces. These collected embeddings, along with face-id and relative-position embeddings, constitute the face embeddings, as detailed in Fig. 7. They are then processed by a Transformer decoder, which predicts distributions over vertex indices at each decoding step, as well as the next-face token and the stopping token. The Transformer's final output layer generates pointer embeddings. These are compared against the vertex embeddings, yielding the final distributions over vertices.

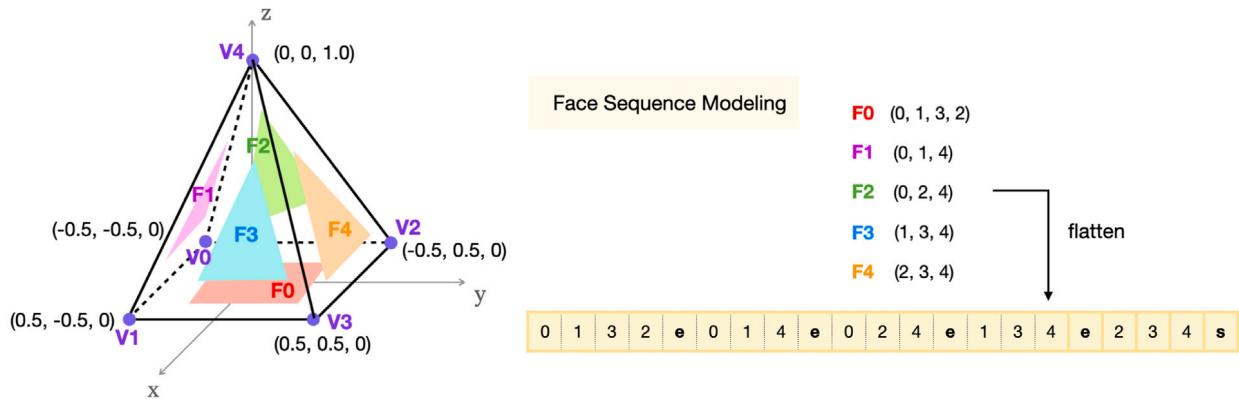

**Fig. 6.** Illustration of creating ordered face sequence. Sorting begins by identifying the smallest vertex index in each face. In situations with identical smallest indices, sorting proceeds by comparing subsequent lower indices within those faces. Vertices in each face are then reordered, starting with the lowest index, and an 'end-face' token $e$ is added to denote the end of each vertex list. The sequence concludes with a stopping token $s$.

### 3.3.1. Nucleus sampling

During inference, we employ Nucleus sampling (Holtzman et al., 2019) as a technique to produce diverse outputs from the Transformer decoder. This method avoids degradation of sample quality by sampling from the top "nucleus" of the probability distribution — a subset with a cumulative probability above a certain threshold, discarding the tail of low probability options.

### 3.3.2. Enforcing constraints for mesh generation

During a single inference, both the vertex and face models are bound by specific rules at the level of the token sequence that ensure the validity of their predictions at each inference step. We thus introduce a series of checks in the form of binary masks to enforce hard constraints present in the mesh data, ruling out predictions that would lead to constraint violations. These checks include the proper placement of stopping tokens, the ordered progression of coordinate and index values, the uniqueness of vertices within faces, etc. If a prediction is discarded, the associated probability mass is redistributed evenly among the remaining valid options. For a detailed description of all constraints and their implementation, please refer to Appendix A.

### 3.3.3. Iterative generation strategy

During mesh generation, when reconstructing a new CAD model from a 3D point cloud, our approach follows a two-stage iterative process: first vertex generation, then face generation, ensuring that each stage adheres to the outlined validity protocols. However, due to the inherent diversity in the sampled predictions, not all semantically valid samples necessarily satisfy the requirements we impose on the CAD models. This issue is common in generative modeling and is typically addressed with some form of aggregation, such as ensembling (Ke et al., 2024). We design an iterative procedure to interleave sampling the solution space with constraint checking – a form of rejection sampling. Importantly, our sampling procedure already implements low-level constraints on the token sequence to enforce the semantic correctness of the modeled sequences, as described in Section 3.3.2. The remaining semantically informed constraints cannot be easily enforced through sampling and thus take the form of hardcoded rules. Consequently, the final CAD model is not generated in a single pass through both stages. Rather, we perform the described checks of intermediate results to enforce the rules and rerun stages when necessary:

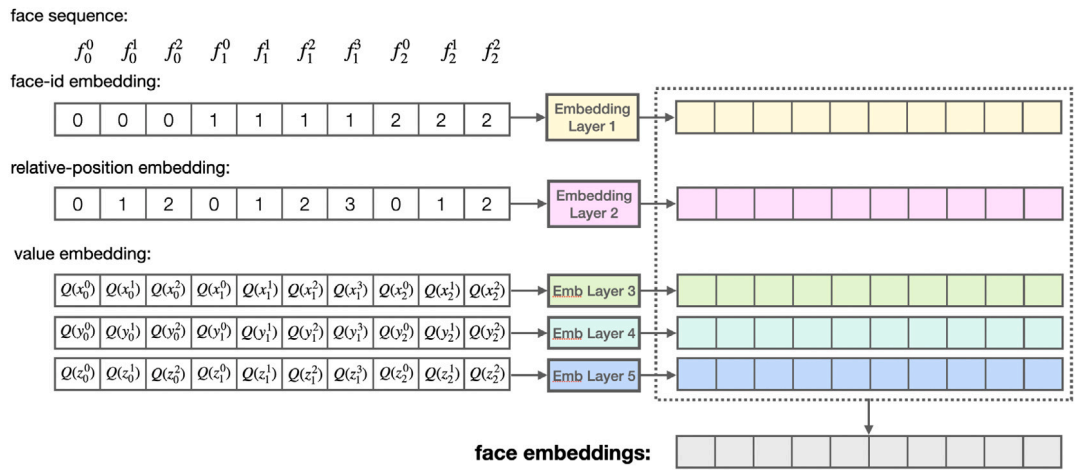

**Fig. 7.** Illustration of the face embedding process, where $f_i^j$ represents the $j$th vertex of the $i$th face in a mesh and $Q(\cdot)$ is a simple quantization function applied to the coordinate values. In this process, the resulting face embeddings for each token are synthesized by integrating three key components: the face-id embedding, the relative-position embedding, and the coordinate value embeddings for the vertices. Specifically, the face-id embedding layer (Embedding Layer 1) accommodates embeddings for up to the maximum number of faces + 1. The relative-position embedding layer (Embedding Layer 2) provides unique embeddings for each position within a face, supporting up to the maximum number of vertices within a face +1. The value embedding layers (Embedding Layers 3, 4, 5) support embeddings for up to a maximum number of all vertices. Each of these embeddings is represented by a vector of 256 dimensions.

Given a 3D point cloud, the algorithm starts by generating a set of vertices during the first stage. This initial step is subject to checking for the presence of the necessary stop tokens. Should the vertices not meet the required criteria, the algorithm regenerates vertices. Upon obtaining a valid set of vertices, the algorithm proceeds to the second stage, faces generation.

Faces are constructed from the vertices using a method that again incorporates validity checks. These checks account for the presence of the stop token and three further criteria that ensure the structural integrity and coherence of the generated CAD model.

(1) The presence of a stop token indicates the completion of the face generation process for the current set of vertices.
(2) The faces that bound the building from below should cover the point cloud projection onto the ground $(x, y)$ plane. Should this coverage be incomplete due to missing vertices or faces, the algorithm responds by initiating a new cycle of vertex generation, undertaking additional iterations of face generation.
(3) The integrity of the building structure is further validated by evaluating the connectivity between the edges of ground polygons and walls. All the edges of the ground polygon(s) should connect appropriately to the vertical edges of the walls. Its failure prompts the regeneration of faces.
(4) Finally, the algorithm checks for the presence of diagonal lines on wall structures. If detected, these trigger additional iterations of face generation.

When the algorithm exhausts the maximum number of iterations for face generation without satisfying all criteria, it takes a step back and regenerates the vertices. This feedback mechanism between the two stages helps model the distribution of CAD models, as it allows us to incorporate the feedback from the face generation step and continue exploring the distribution of vertices before arriving at the final valid CAD model of the building. The algorithm is described in Alg. 1 in Appendix B.

## 4. Experiments

### 4.1. Dataset

**3D point clouds.** Our main dataset comes from Zurich, Switzerland. The LiDAR point clouds used for the building reconstruction come from the Swiss Federal Office of Topography (swisstopo, 2023). They include semantic category labels and are provided in LAS file format, with the points organized into square tiles, each covering an area of 1 km$^2$. Following a modified version of the classification guidelines (Yan et al., 2015) established by the American Society for Photogrammetry and Remote Sensing (ASPRS), the points have been classified into six distinct classes: unclassified and temporary objects; ground; vegetation (further divided into low, medium, and high); buildings; water; bridges and viaducts. For our purpose, we exclusively utilize points labeled as buildings.

**Polygonal meshes.** The dataset comprising the corresponding polygonal mesh models comes from the open data released by the City of Zurich (Stadt Zurich, 2023). It includes 3D representations of all current buildings within the city, crafted to Level of Detail 2 (LoD2). LoD2 models offer a refined depiction of buildings, including the precise geometry of roof structures and other major architectural features, but not minor facade details or interior structures. The data format varies, encompassing Drawing Interchange Format (.dxf), geodatabases (.gdb), GeoPackage (.gpkg), and others. For our analysis, we have processed all buildings into discrete polygonal meshes. Each building model is represented as a collection of vertices and flat faces that connect the vertices to form the building's exterior surface. Example meshes are shown in Fig. 8.

Overall, the area used for our project, where the two data sources overlap, spans 9 km by 8 km. We partition that region into spatially disjoint training and test sets as depicted in Fig. 9.

Our work focuses on the learning-based reconstruction of 3D digital models from point clouds of individual buildings. Recognizing the complexity of segmenting the city-scale point cloud into individual buildings, we simplify the preprocessing of the dataset by sidestepping the segmentation problem. Instead, we first use the classification labels in the LiDAR scans to filter the data, then partition the points into subsets associated with separate buildings with the help of the reference footprints. Exemplary per-building point clouds are visualized in Fig. 10.

**Additional datasets.** We have also tested our method on datasets collected in two more cities: Berlin, Germany, and Tallinn, Estonia. The dataset from Berlin covers an area of 4 km × 4 km, comprising 69,132 buildings for training and 7,808 for testing. The source of both point clouds and corresponding polygonal meshes is the Open Data Initiative of the state of Berlin, available at Stadt Berlin (2021) and Berlin Business Location Center (2015). The Tallinn dataset spans a 3 km × 3 km area, containing 33,397 buildings for training and 2,670 for testing. The LiDAR point clouds and meshes are both supplied by the Estonian Land Board for Spatial Data (Estonian Land Board, 2021). Examples are shown in Fig. 11.

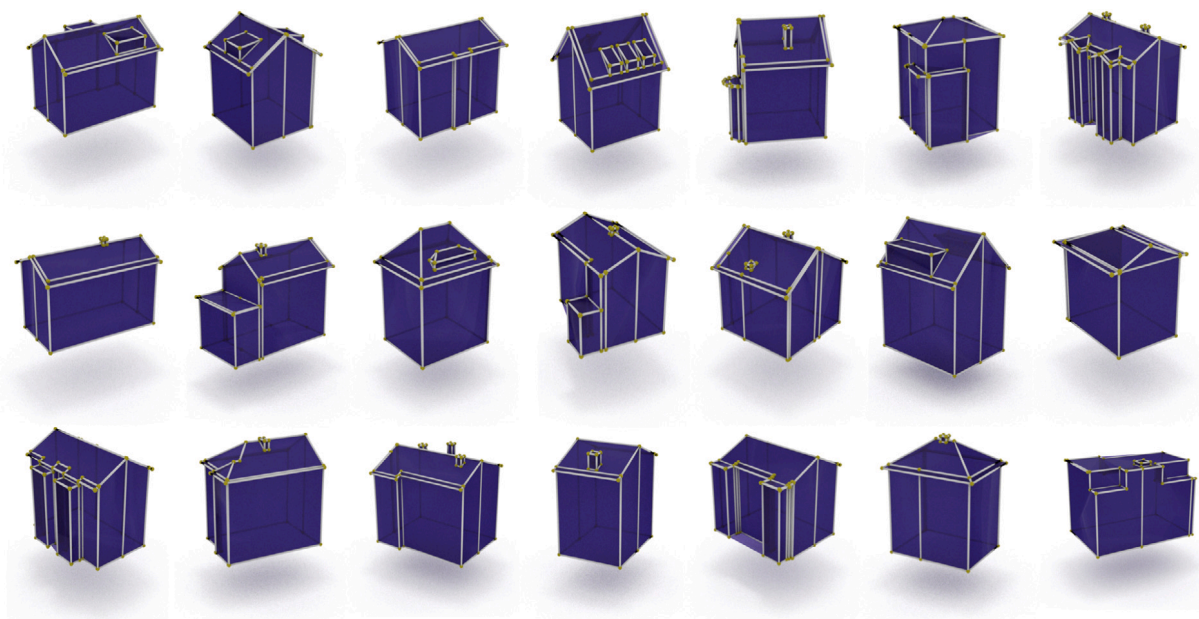

**Fig. 8.** Examples of polygonal building meshes from the Zurich dataset.

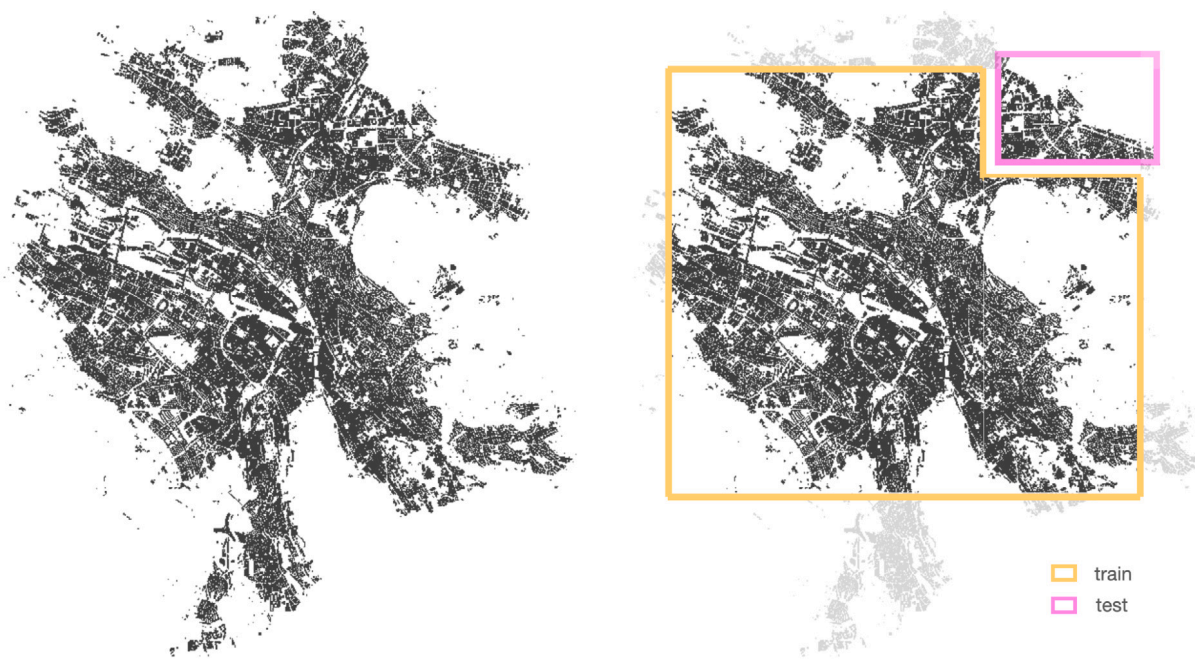

**Fig. 9.** Zurich city buildings (left) and the corresponding train-test split dataset used in experiments (right).

### 4.2. Implementation details

We exclude buildings with more than 100 vertices or over 500 faces to maintain reasonable computation complexity during training and inference. Moreover, we limit the processing to 10 resampling iterations in Alg. 1.

During training, we employ data augmentation to increase robustness. Specifically, we apply random scaling along the $x$, $y$, and $z$ axes by random factors within [0.8, 1.2]. We further augment by rotating around the vertical ($z$) axis by random angles. Moreover, we also apply slight perturbations to the vertex locations, with each vertex shifted by up to 10% of the object diameter. During testing, we use nucleus sampling. The main hyper-parameter of that method, top-$p$, restricts sampling to the smallest possible set of tokens that cover a cumulative probability mass of $p$. We empirically fix it to 0.9, the default value proposed for PolyGen.

We trained the vertex and face modules for 500,000 weight updates each, using the Adam optimizer with a batch size of 16, and with a gradient clipping at a maximum norm of 1.0 for stability on a single NVIDIA GeForce RTX 3090 Ti GPU. We adopt a cosine annealing scheme for the learning rate schedule, commencing from a maximum learning rate of $3 \times 10^{-4}$, while also applying a linear warm-up during the first 5000 steps. Dropout at a rate of 0.2 was applied in training both

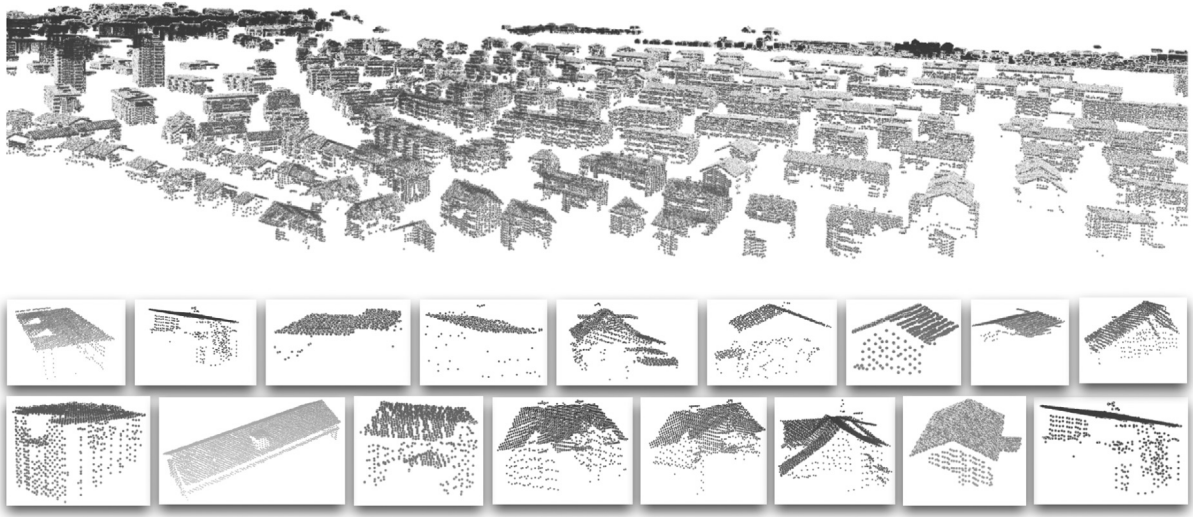

**Fig. 10.** Point cloud samples from the Zurich dataset, featuring an overview at the top and detailed zoom-ins of individual building point clouds at the bottom.

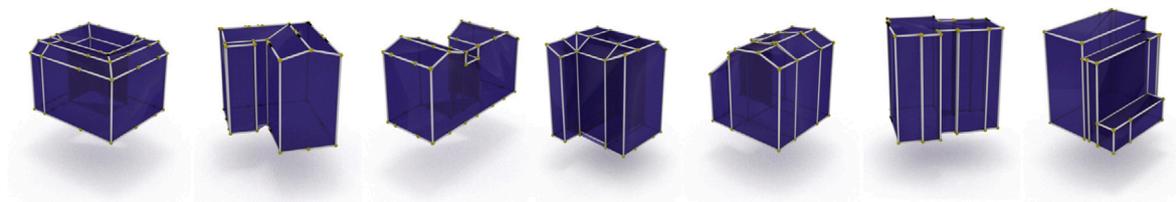

(a) Samples of building meshes in Berlin dataset

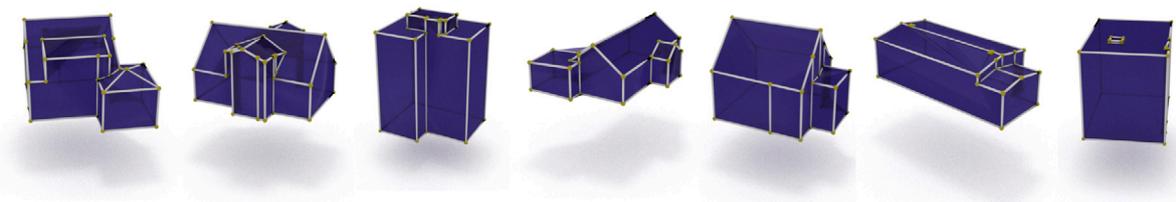

(b) Samples of building meshes in Tallinn dataset

**Fig. 11.** Examples of polygonal building meshes from the Berlin and Tallinn datasets.

modules. Training took 44 hours for the vertex module, respectively 29 hours for the face module.

### 4.3. Evaluation metrics

We utilize three different distance metrics to assess the fidelity of our reconstructed CAD models.

**Mean Distance Error** (MDE) serves as a metric to quantify the average deviation from the ground truth w.r.t. both vertices and surfaces. To compute the MDE on surfaces, we uniformly sample 10,000 points on the surface of the ground truth mesh and project them onto the nearest surface plane of the predicted mesh to obtain a set of point correspondences. The MDE is calculated as the mean Euclidean distance between corresponding points, measuring the average spatial distance between the true and predicted surfaces.

**Hausdorff distance** is used to measure the maximum extent of deviation between the reconstructed surface and the ground truth mesh. It captures the maximum distance from a point on one surface to the closest point on the other, thus providing the worst-case scenario regarding per-point error. The equation for the Hausdorff distance, where $A$ and $B$ represent the uniformly sampled 10,000 point sets from the ground truth and reconstructed surfaces respectively, is given by:

$$H(A, B) = \max \left\{ \sup_{a \in A} \left( \inf_{b \in B} d(a, b) \right), \sup_{b \in B} \left( \inf_{a \in A} d(a, b) \right) \right\},$$

with $d(\cdot, \cdot)$ the Euclidean distance.

**Chamfer distance** is also calculated to evaluate the average closest point distance between vertex sets and surfaces. This metric measures the minimum deformation needed to "warp" the prediction to the ground truth and gives a sense of the overall shape difference. The distance computation for surfaces follows the same sampling strategy as above. The Chamfer distance is defined as follows:

$$C(A, B) = \frac{1}{|A|} \sum_{a \in A} \sum_{b \in B} \min d(a, b) + \frac{1}{|B|} \sum_{b \in B} \sum_{a \in A} \min d(b, a).$$

**Precision, Recall and F1-score for vertices and edges.** To further evaluate the performance of vertex and edge reconstruction within our model, we measure precision, recall, and F1-score for vertices and edges. For vertex pairs, we calculate the Euclidean distance between corresponding vertices in the predicted and ground truth models. A vertex is considered correctly reconstructed if this distance is below a

**Table 1**

Evaluation on vertex prediction quality with different methods: City3D (Huang et al., 2022), 2.5D Dual Contour (Zhou and Neumann, 2010) and Ours. Best values are bold.

|  | Precision ↑ | Recall ↑ | F1-score ↑ | Chamfer distance ↓ |
|---|---|---|---|---|
| City3D | 0.6717 | 0.7364 | 0.6804 | 0.9387 |
| 2.5D Dual Contour | 0.2899 | 0.5060 | 0.3553 | 1.4006 |
| Our Vertex-module | **0.8928** | **0.8617** | **0.8677** | **0.5816** |

**Table 2**

Various errors (m) of reconstruction with different methods: City3D (Huang et al., 2022), 2.5D Dual Contour (Zhou and Neumann, 2010) and Ours. Best values are bold.

|  | MDE ↓ | Hausdorff distance ↓ | Chamfer distance ↓ |
|---|---|---|---|
| City3D | 0.3046 | 1.4723 | 0.3708 |
| 2.5D DualContour | 0.3646 | 1.9157 | 0.4368 |
| Ours | **0.2542** | **1.1200** | **0.3060** |

**Table 3**

Ablation studies on the necessity of all components.

|  | MDE ↓ | Hausdorff distance ↓ | Chamfer distance ↓ |
|---|---|---|---|
| Ours: alternative vertex | 0.1885 | 0.9695 | 0.2337 |
| Ours: w/o constraints | 0.1624 | 0.7521 | 0.2197 |
| Ours: single inference | 0.1998 | 1.2584 | 0.2554 |
| Ours: w/o data augmentation | 0.2542 | 1.1296 | 0.3060 |
| Ours: full | **0.1554** | **0.7488** | **0.2149** |

preset threshold. For edge pairs, we first sample a fixed, equal number of points along each edge (100 points in our experiments) and compute the mean (squared) distance between the two point sets. Based on this distance and a preset threshold, we then calculate the precision, recall, and F1-score. These metrics provide a measure of reconstruction accuracy for the individual mesh elements by counting the fraction of vertices or edges that have been recovered within a set tolerance. The three metrics are computed under various distance thresholds to evaluate how the model performance varies with different vertex and edge alignment tolerances.

*4.4. Quantitative results*

We compare our method against two representative baselines, City3D (Huang et al., 2022) and the 2.5D Dual Contouring method (Zhou and Neumann, 2010).

First, we assess the accuracy of vertex prediction. The precision, recall, and F1-score metrics indicate a high degree of correctness in identifying vertices. With a chosen true positive threshold set at 1 m, deemed reasonable for the accuracy requirements of building reconstruction, these metrics are reported in Table 1. We also plot the F1-score curve against varying thresholds in Fig. 13(a). Our approach maintains a higher score across different conditions, meaning it consistently retrieves vertices and edges more correctly across a range of geometric accuracy requirements. Further, the lower Chamfer distance confirms the better precision of the reconstructed vertex locations, meaning that the building surfaces reconstructed by our method are geometrically closer to the ground truth shapes. We also examine the absolute error in the predicted number of vertices relative to the ground truth. Compared to other methods, the histogram Fig. 12(a) reveals that a larger fraction of all samples have small errors, implying that our method gets closest to the true vertex counts.

Next, we assess the absolute errors in the lengths of edges and the areas of faces, as illustrated in Fig. 12(c) and Fig. 12(d), respectively. The errors with our method are more concentrated near zero for both metrics, indicating that it most precisely reconstructs edge and face dimensions.

Furthermore, the F1-score of the proposed method remains higher than those of the two baselines across a wide range of tolerance values, Fig. 13(b), meaning that, at any given accuracy threshold, it achieves a better balance between recall and precision.

Finally, in terms of surface quality metrics, Table 2, our method achieves the lowest mean distance error, Hausdorff distance, and Chamfer distance w.r.t. the ground truth models. These metrics collectively emphasize the fidelity of our reconstruction, with the mean distance highlighting average deviation, the Hausdorff distance indicating maximum discrepancy, and the Chamfer distance reflecting overall shape similarity.

*4.5. Qualitative results*

When visually inspecting the reconstructed building models, as depicted in Fig. 14, our method effectively deals with incomplete input point clouds, characterized by denser points on roofs and sparser points on walls. The generative model can reconstruct fine structural details, such as thin chimneys, even from minimal evidence. In contrast, alternative methods often yield overly simplified models that omit such details or create models with excessive spurious edges. We also present a gallery of perspective views with many reconstructed building blocks in Fig. 15, comparing the reconstruction quality of different approaches. Further results for the entire test region of Zurich are shown in Fig. 17.

As a particular feature, our dataset includes numerous houses with roof overhangs — a feature that existing methods typically struggle to replicate. The sequential polygon prediction in our generative model successfully captures these overhangs, as shown in Fig. 15 (bottom), underscoring its ability to handle varied mesh geometries when supervised with appropriate reference data.

**Failure cases.** Fig. 16 shows instances where our models did not yield the desired results. These failures usually fall into two categories. The first one is erroneous vertex predictions: omitted or misplaced vertices inevitably lead to significant deviations from the ground truth since vertices are the geometric basis for the mesh faces. The other main failure mode is inadequate face generation. Unsurprisingly, the model performance tends to decline with increasingly complex building geometry. The generation of faces proves particularly challenging for intricate roof designs that result in a scarcity of input points on finer sub-structures.

*4.6. Ablation study*

**Vertex Module.** We investigate the importance of sequential, autoregressive prediction for the first ablation study. To do so, we designed an alternative method for vertex generation. The point cloud encoder stays the same, but instead, the sequential prediction is followed by a decoder that predicts a set of *unordered* 3D vertex coordinates, also via a series of Transformer blocks. The inputs to the decoder are the point cloud features and a set of query embeddings. The decoder transforms those inputs into a new set of features. From those features, a prediction head in the form of a three-layer perceptron extracts 3D vertex coordinates, and another prediction head extracts classification scores that indicate the likelihood of each vertex being present in the reconstruction. The unordered model is supervised with an L2 loss for the coordinates and a binary cross-entropy loss for the classification scores. The set of vertices to be passed on to the subsequent face module is determined by thresholding the classification score at 0.5. Results are presented in Table 3, labeled as "Ours: alternative vertex". They indicate a decline across all quality metrics, suggesting that the autoregressive design of the original generative model is crucial to adequately capturing the spatial relationships between vertices and is more conducive to the subsequent mesh face reconstruction.

**Token-level constraints.** Our method integrates several constraints that guide the generation of vertices and faces, ensuring the resulting 3D models adhere to specified geometric rules and architectural principles. To evaluate the contribution of these token-level constraints to the quality of the reconstructed models, we conducted an ablation study where we removed the plausibility checks from the shape generation process. When not bound by the constraints, the transformer network operates with greater freedom, providing insights into its intrinsic capabilities and the underlying data representation. Without constraints, the generation pipeline IS solely dependent on the patterns learned from the training data, without additional guidance to refine the placement of vertices or the construction of faces. The results are shown in Table 3 as "Ours: w/o constraints". We observe a decrease across all quality metrics upon removing the token-level constraints, indicating that although our network has extracted useful patterns from the training data, the learned patterns do not constrain it tightly enough.

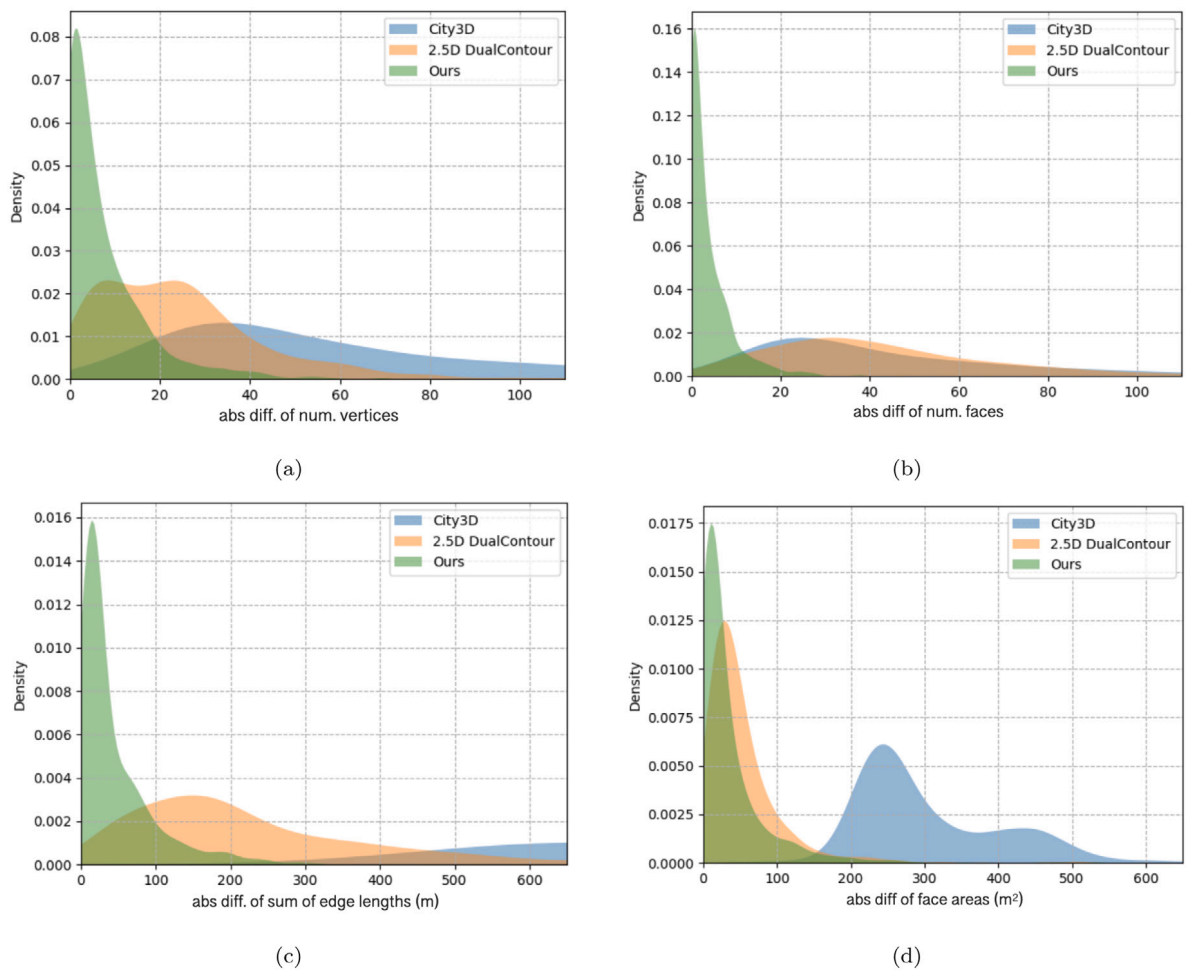

**Fig. 12.** Error distributions with different reconstruction methods. (a) Absolute difference of vertex counts. (b) Absolute difference of face counts. (c) Absolute difference of the sum of edge lengths in meters. (d) Absolute difference of face areas in square meters.

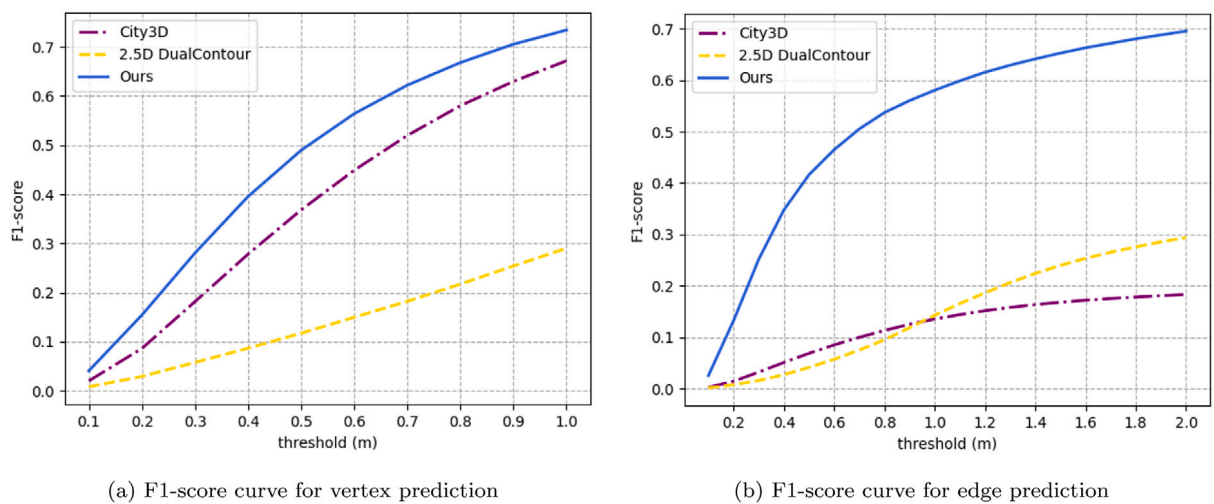

(a) F1-score curve for vertex prediction

(b) F1-score curve for edge prediction

**Fig. 13.** F1-score curves for vertex and edge prediction with different algorithms.

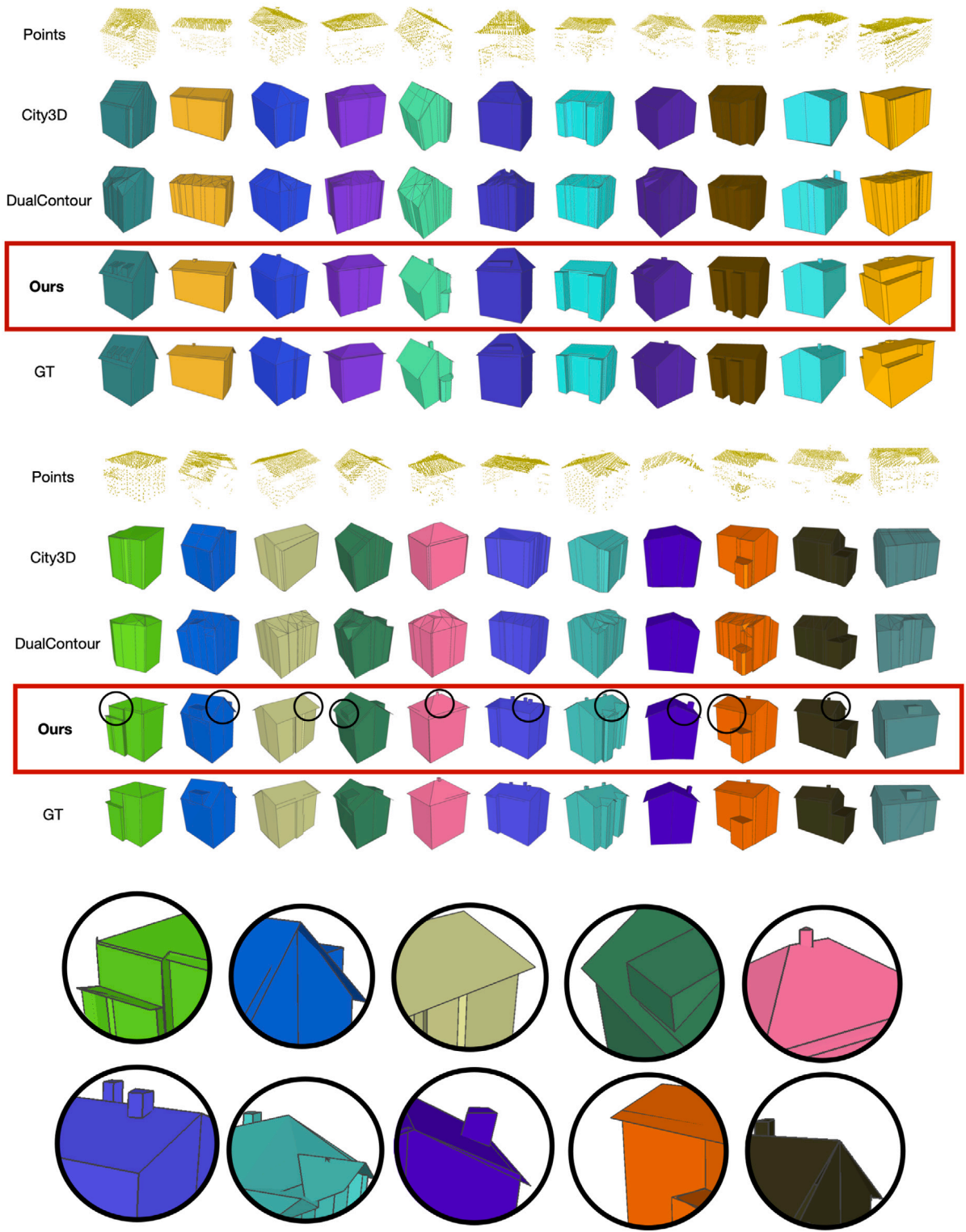

**Fig. 14.** Gallery of building reconstructions from different methods. Detailed views at the bottom highlight the ability of our method to reconstruct detailed features such as roof overhangs and chimneys.

**Single inference.** We also conducted an ablation study in which we disabled the iterative generation that is part of our model by default. The iteration scheme is designed to repeatedly generate complete building hypotheses until the output fulfills the predefined set of rules, ensuring the validity of the predicted model. To understand the significance of that procedure, we modified it to generate a single result for each test sample, with the token-level constraints enabled, but without regard for the validity of the predicted full building model. The corresponding quality metrics are shown in Table 3 as "Ours: single inference". They point to a marked decline in performance if we forgo the iterative generation and limit ourselves to single-shot generation per sample, irrespective of semantic plausibility. This finding highlights the importance of the quality filter and associated iteration scheme in refining the reconstruction iteratively.

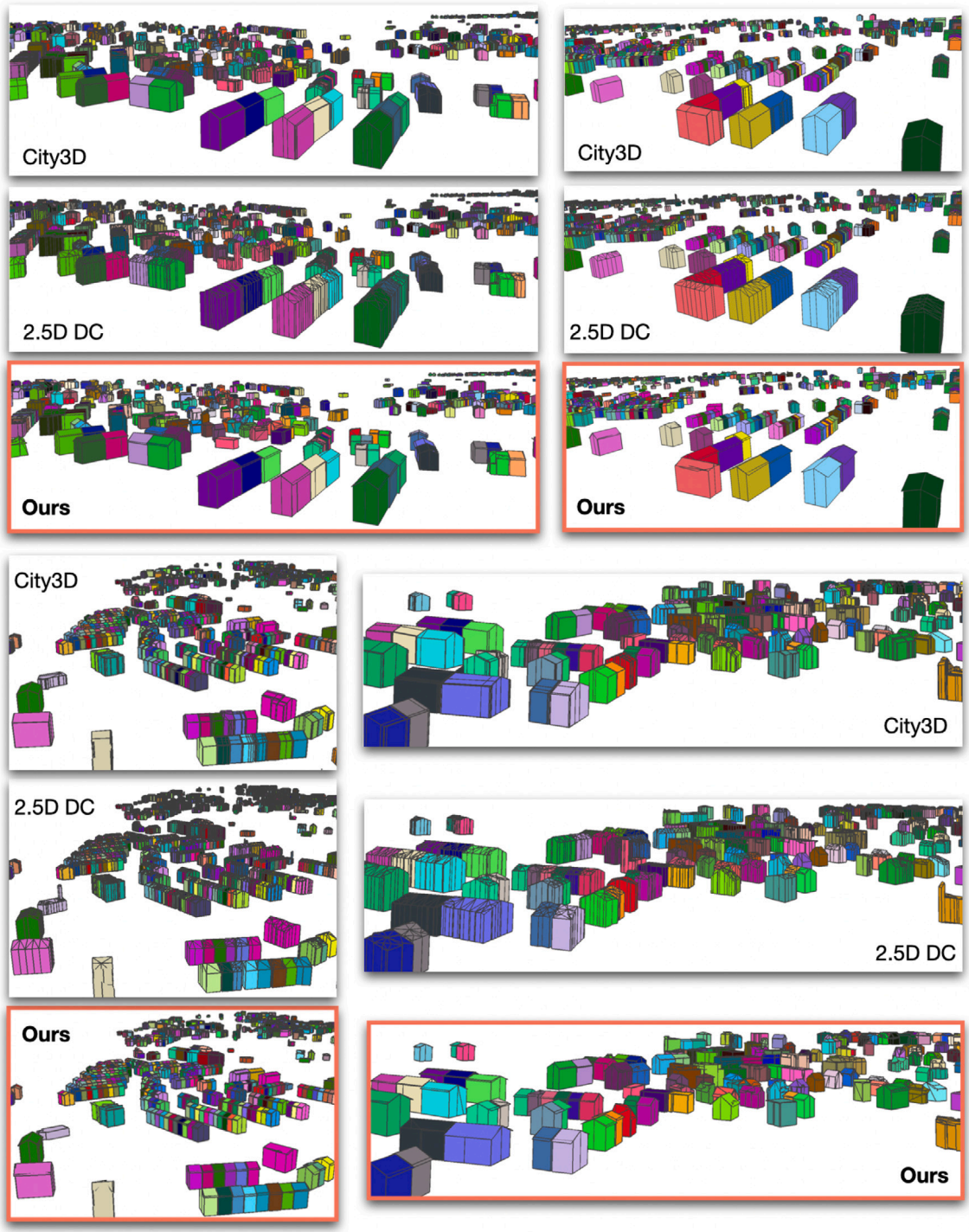

**Fig. 15.** Gallery of reconstructed building blocks. Each vertical triplet of images showcases the reconstruction results from City3D, 2.5D DualContour, and our method, respectively.

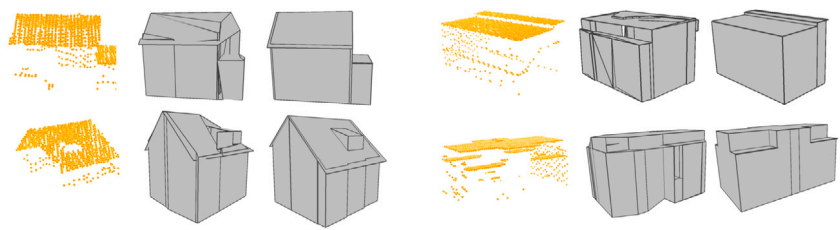

**Fig. 16.** Examples (input, reconstruction, and reference) with large reconstruction errors.

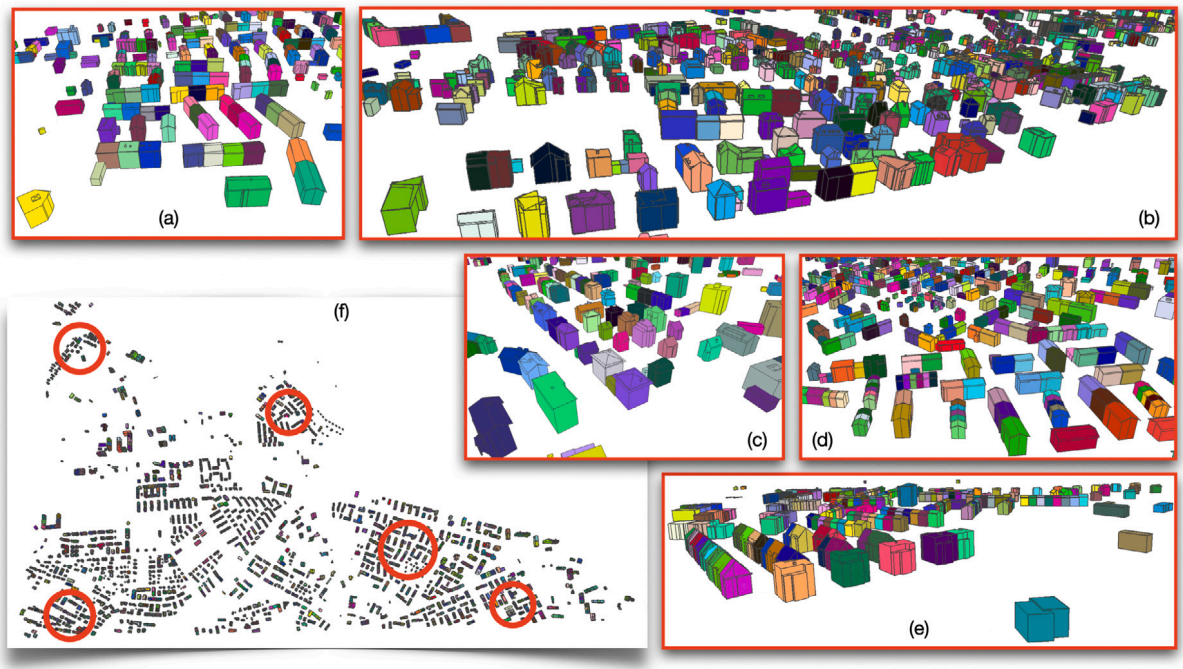

**Fig. 17.** Reconstructed building blocks in Zurich. (f): Reconstruction of a 3 km × 2 km area in Zurich. (a) – (e) Magnified views of selected regions.

**Data augmentation.** During the training phase, data augmentation techniques, specifically scaling, rotation around the *z*-axis, and adding noise, were employed. An ablation study was conducted to evaluate their impact. The results, as detailed in Table 3 ("Ours: w/o data augmentation"), show a significant quality gain attributable to data augmentation. Thus, our model is likely not trained to its full performance and would benefit from even larger quantities of training data, especially given the favorable scaling properties of transformer architectures.

*4.7. Results of reconstructing other cities*

Beyond Zurich, we have also applied our method to LiDAR data from Berlin and Tallinn, retraining the models for each city's datasets. The results are depicted in Figs. 18 and 19. Despite the variations in building shapes between different cities, the reconstruction outcomes consistently exhibit high geometric quality.

*4.8. Discussion*

Several challenges remain when handling complex roofs and fine architectural elements, particularly if the associated point clouds are relatively sparse. Future efforts could explore ways to enhance the model's sensitivity to small roof structures, potentially through more powerful point cloud encoders or architectural innovations promoting even better contextual reasoning. While our method aims to produce outputs that resemble the ground truth data, it does not explicitly enforce geometric properties like watertightness or manifoldness. This reflects the characteristics of our training data, in which many models are not watertight, especially in the presence of overhanging roofs. The reference data suggest that watertightness is also not always required. If it is needed, it can be achieved with appropriate post-processing. A conceptual limitation of our current approach is the independent prediction modules for vertices and faces, which may hamper the synergies between vertex and face processing. Future extensions should seek to integrate the two subtasks more tightly, which could potentially improve reconstruction accuracy while reducing the computational cost of inference. Finally, we have concentrated on the reconstruction step. For a fully automatic pipeline, future research will be needed to segment the raw scan data into individual building units reliably. That step, a prerequisite for a complete, highly automated urban modeling tool, can draw from the rich literature on detection and segmentation in 3D point clouds.

**5. Conclusion**

We have studied the task of automatically reconstructing 3D polygonal mesh models of buildings from airborne LiDAR point clouds, and have proposed an autoregressive, generative model based on the transformer architecture. The model consists of two parts, a vertex module for sequential vertex generation and a subsequent face module to that connects vertices to mesh faces. When combined with geometric plausibility checks and rejection sampling, the novel model consistently outperforms both a classical optimization-based and a recent learning-based baseline across a range of performance metrics.

**CRediT authorship contribution statement**

**Yujia Liu:** Conceptualization, Data curation, Methodology, Validation, Writing – original draft, Writing – review & editing. **Anton Obukhov:** Conceptualization, Methodology, Supervision, Validation, Writing – original draft, Writing – review & editing. **Jan Dirk Wegner:** Conceptualization, Funding acquisition, Project administration, Supervision, Writing – original draft, Writing – review & editing. **Konrad Schindler:** Conceptualization, Funding acquisition, Methodology, Supervision, Validation, Writing – original draft, Writing – review & editing.

**Declaration of competing interest**

The authors declare that they have no known competing financial interests or personal relationships that could have appeared to influence the work reported in this paper.

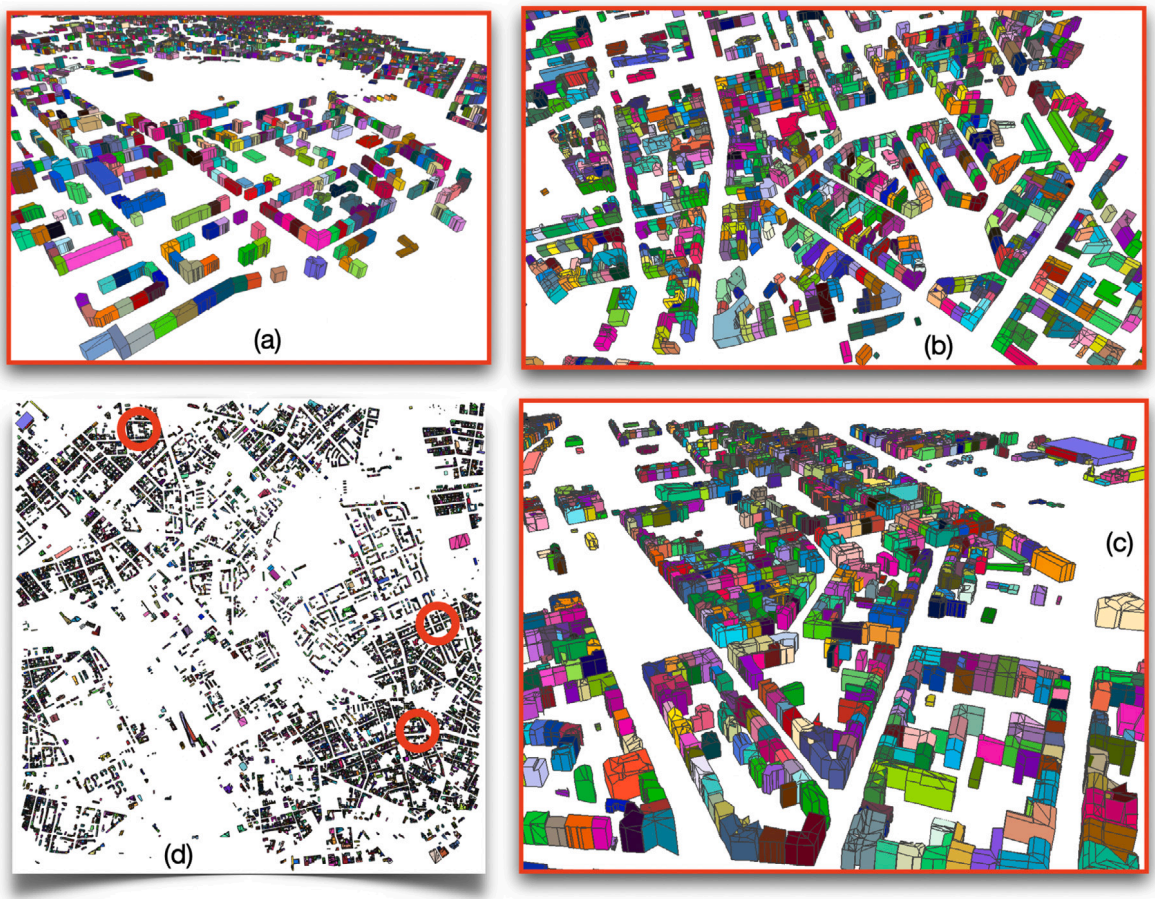

**Fig. 18.** Reconstructed building blocks in Berlin. (d) Reconstruction of a 4 km × 4 km area in Berlin. (a) – (c) Magnified views of selected regions.

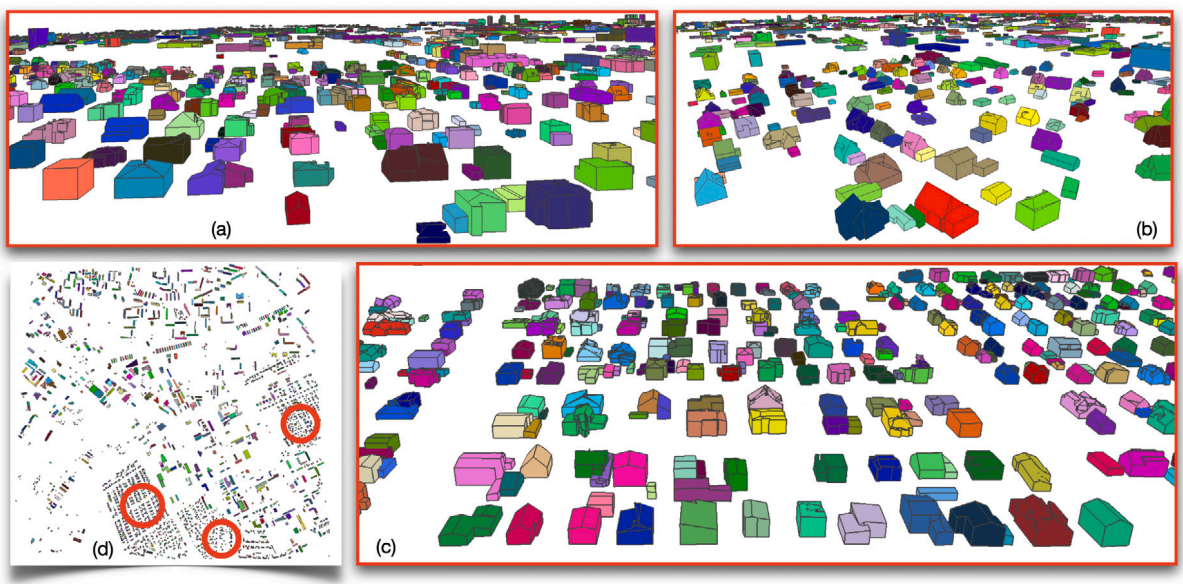

**Fig. 19.** Reconstructed building blocks in Tallinn. (d) Reconstruction of a 3 km × 3 km area in Tallinn. (a) – (c) Magnified views of selected regions.

## Appendix A. Constraints for mesh generation

For the vertex module:

(1) Stopping Token Placement: The sequence must place the stopping token only after an $x$-coordinate.
(2) $z$-Coordinate: The sequence of sampled $z$-coordinates must be non-decreasing.
(3) $y$-Coordinate: The subsequent $y$-coordinates should not decrease, if their corresponding $z$-coordinates are equal.
(4) $x$-Coordinate: The subsequent $x$-coordinates must strictly increase when their $y$ and $z$ counterparts are equal to those preceding.

For the face module:

(1) Face Index Progression: The first generated index of a new face must be greater or equal to the first index of the previous face.
(2) Vertex Index Order: Within a face, subsequent vertex indices must be greater than the first index.
(3) Vertex Uniqueness: All vertex indices within a face are required to be distinct.

## Appendix B. Algorithm for reconstruction and validation of a polygonal mesh

**Algorithm 1** Reconstruction and Validation of a Polygonal Mesh

1: **function** RECONSTRUCTMESH(*pointCloud*, *maxIterationsVertices*, *maxIterationsFaces*)
2:     $iterationVertices \leftarrow 0$
3:     **while** $iterationVertices < maxIterationsVertices$ **do**
4:         $vertices \leftarrow$ GENERATEVERTICESFROMPOINTCLOUD(*pointCloud*)
5:         **if not** HAVE_STOP_TOKEN(*vertices*) **then**
6:             $iterationVertices \leftarrow iterationVertices + 1$
7:             **continue**
8:         $iterationFaces \leftarrow 0$
9:         **while** $iterationFaces < maxIterationsFaces$ **do**
10:             $faces \leftarrow$ GENERATEFACESFROMVERTICES(*vertices*)
11:             **if not** HAVE_STOP_TOKEN(*faces*) **then**
12:                 $iterationFaces \leftarrow iterationFaces + 1$
13:                 **continue**
14:             **if not** IS_FLOOR_COVERING_POINTCLOUDXY(*faces*) **then**
15:                 **if** ARE_MISSING_FLOOR_VERTICES(*faces*) **then**
16:                     **break**
17:                 **else if** ARE_MISSING_FLOOR_FACES(*faces*) **then**
18:                     $iterationFaces \leftarrow iterationFaces + 1$
19:                     **continue**
20:             **if not** ARE_FLOOR_EDGES_CONNECTED_TO_WALLS(*faces*) **then**
21:                 $iterationFaces \leftarrow iterationFaces + 1$
22:                 **continue**
23:             **if** ARE_DIAGONAL_EDGES_PRESENT_ON_WALLS(*faces*) **then**
24:                 $iterationFaces \leftarrow iterationFaces + 1$
25:                 **continue**
26:             $cadModel \leftarrow$ BUILDCADMODEL(*vertices*, *faces*)
27:             **return** $cadModel$
28:         $iterationVertices \leftarrow iterationVertices + 1$
29:     **raise** "Maximum number of iterations has been reached"